\documentclass{article}
\usepackage[markup=underlined]{changes}
\usepackage{spconf,amsmath,amsfonts,amssymb,amsthm,graphicx,tabularx}
\usepackage{pgfplots}
\usetikzlibrary{spy}

\newcommand\bbR{{\mathbb R}}
\newcommand\bbE{{\mathbb E}}
\newcommand\bfx{{\mathbf x}}

\usepackage{hyperref}

\title{LEARNING PARTIAL DIFFERENTIAL EQUATIONS FROM DATA USING NEURAL NETWORKS}
\name{Ali Hasan$^{\dagger}$ \qquad Jo\~{a}o M. Pereira$^{\star}$ \qquad Robert Ravier$^{\star}$ \qquad Sina Farsiu$^{\dagger\star}$ \qquad Vahid Tarokh$^{\star}$\thanks{ This research is funded in part by DARPA grant No. HR00111890040.
}}

			\address{$^{\dagger}$ Department of Biomedical Engineering, Duke University, Durham NC 27708 \\
			    $^{\star}$ Department of Electrical and Computer Engineering, Duke University, Durham NC 27708}
\begin{document}
%\ninept
%
\maketitle
\begin{abstract}
We develop a framework for estimating unknown partial differential equations (PDEs) from noisy data, using a deep learning approach. 
Given noisy samples of a solution to an unknown PDE, our method interpolates the samples using a neural network, and extracts the PDE by equating derivatives of the neural network approximation.
Our method applies to PDEs which are linear combinations of user-defined dictionary functions, and generalizes previous methods that only consider parabolic PDEs. 
We introduce a regularization scheme that prevents the function approximation from overfitting the data and forces it to be a solution of the underlying PDE.
We validate the model on simulated data generated by the known PDEs and added Gaussian noise, and we study our method under different levels of noise.
We also compare the error of our method with a Cramer-Rao lower bound for an ordinary differential equation (ODE).
Our results indicate that our method outperforms other methods in estimating PDEs, especially in the low signal-to-noise (SNR) regime.
\end{abstract}
\begin{keywords}
Partial differential equations, neural networks, Cramer Rao bound.
\end{keywords}
\section{Introduction}

\label{sec:intro}
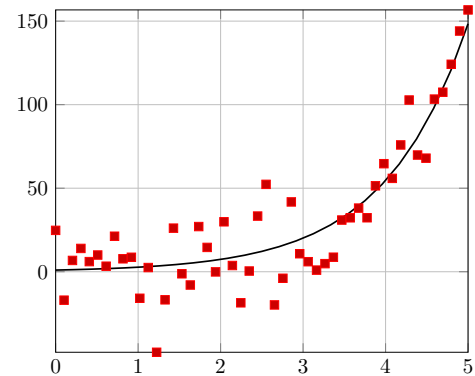
\begin{figure}
    \centering
\scalebox{0.8}{
\begin{tikzpicture}[spy using outlines={rectangle, magnification=3,connect spies}]
\begin{axis}[grid=major,domain=0:5,enlargelimits=false]
\addplot [no marks,thick]{exp(x)};
\addplot+[ only marks]table[x=x,y=y,col sep=comma]
{test_data.csv};
\coordinate (spypoint) at (axis cs:2,0.2);
\coordinate (spyviewer) at (axis cs:0,90);
%\spy[width=3cm,height=3cm] on (spypoint) in node [fill=white] at (spyviewer);
\end{axis}
\end{tikzpicture}
}
    \caption{Schematic of training process. The black line is the function trained by the neural network. $\mathcal{L}_u$ forces this function to match the data and $\mathcal{L}_d$ forces it to be smooth and recovers the equation}
    \label{fig:schematic}
\end{figure}

Partial differential equations (PDEs) are widely used in several quantitative disciplines, from physics to economics, in an attempt to describe the evolution and dynamics of phenomena. 
Often, such equations are derived using first principle approaches in conjunction with data observations. 
However, as datasets are often cumbersome and difficult to analyze through traditional means, there arises a need for the automation of such processes. 
We present an approach that uses neural function approximation with function regularization to recover the governing partial differential equations a wide range of possible PDEs.

Recent work \cite{rudy2017data, brunton2016discovering} used finite difference and polynomial approximation to extract the governing equation from data. 
However, the methods perform poorly in the presence of noise.
In \cite{long2017pde}, the authors use convolutional neural networks with filters constrained to finite difference approximations to learn the form of a PDE, but no sparsity constraint is enforced on the learned PDE, which may lead to verbose solutions.
Although the approaches described in \cite{berg2019data} and \cite{xu2019dl} consider deep learning approaches for function approximation, these do not analyze the low signal-to-noise (SNR) regime. 
All previous methods \cite{rudy2017data,berg2019data,xu2019dl} consider parabolic PDEs, that is, PDEs that depend on first order derivatives of the function in time. Our work is, to the best of our knowledge, the first to consider more general PDEs.

Similar to previous methods \cite{rudy2017data,berg2019data,xu2019dl}, we consider PDEs which are linear combinations of dictionary functions, but instead of assuming this linear combination has a dependence on time, we recover the underlying PDE from the null space of the dictionary functions.
We interpolate the observed data using a neural network, and include a regularization term that forces the neural network to follow the PDE that best describes the data.
In order to calculate this regularization term, we use automatic differentiation to calculate derivatives of the neural network.
We finally determine the underlying PDE by checking the learned regularization term.
A byproduct of this dual optimization is the increase in performance of the method in the low SNR regime.
To summarize, our main contributions are:
\begin{enumerate}
    \item Establish a deep learning framework for the identification of PDEs which are linear combinations of user-defined dictionary functions. \vspace{-.2cm}
    \item Introduce a regularization scheme for preventing function approximators from overfitting to noise. \vspace{-.2cm}
    \item Compare our method with the Cramer-Rao lower bound for a simple ordinary differential equation (ODE).
\end{enumerate}

Our paper is organized as follows. In Section \ref{sec:problem} we introduce the PDE estimation problem. In Section \ref{sec:methods}, we present our approach to estimating the underlying PDE using neural networks. In Section \ref{sec:cramerrao} we present a Cramer-Rao lower bound \cite{cramer1946mathematical} for a simple ODE and compare it with the results obtained by our methods. Finally, in Section \ref{sec:experiments} we present numerical results obtained with simulated data and in section \ref{sec:majhead} we summarize our contributions and discuss future applications.

\section{Problem}
\label{sec:problem}

A PDE is a equation that relates a function with its partial derivatives. For a function $u: \Omega \to \bbR$, where $\Omega \subset \bbR^N$, and a vector of non-negative integers $\alpha=(\alpha_1,\dots,\alpha_n)$, we denote $D^\alpha u := \frac{\partial^{\alpha_1}}{\partial x_1^{\alpha_1}}\cdots\frac{\partial^{\alpha_n}}{\partial x_n^{\alpha_n}} u$ and $|\alpha| = \alpha_1 + \dots + \alpha_n$. Using this notation, we can define any PDE by
 \vspace{-.08cm}
\begin{equation}
F(\bfx,u(\bfx),D^\alpha u (\bfx)) = 0,
\label{eqn:law}  \vspace{-.08cm}
\end{equation}
where $F$ is a function that relates $\bfx\in \Omega$ with $u$ and its partial derivatives at $\bfx$. We say $u$ is a solution to the PDE if \eqref{eqn:law} holds for every point $\bfx\in\Omega$. In this paper, we observe data points $(\bfx,y) =  \{(\mathbf{x}_1,y_1) \dots  , (\mathbf{x}_J,y_J)\}$, where $y$ are noisy function values at $\bfx$, that is,
 \vspace{-.08cm}
\begin{equation}
y_i = u(\mathbf{x}_i) + \xi_i, \quad i=1,\dots,J,
\label{eq:noisydatadefinition}   \vspace{-.08cm}
\end{equation}
where the noise $\xi_i,\,i=1,\dots,J$ is assumed to be Gaussian, with variance $\sigma^2$, and $u$ is a solution to \eqref{eqn:law} for some unknown function $F$. Since the space of possible functions is prohibitively large, we restrict our attention to PDEs that are linear combinations of dictionary functions, relating $u$ and its derivatives. More formally, let $\mathcal{D}=\{\mathcal{D}_1,\dots, \mathcal{D}_L\}$ be such a dictionary, we consider PDEs of the form
 \vspace{-.08cm}
\begin{equation}
\sum_{i=1}^L \phi^*_i \mathcal{D}_i (\bfx,u(\bfx),D^\alpha u (\bfx)) = 0,\quad \forall \bfx\in \Omega,  \vspace{-.08cm}
\label{eqn:law2}
\end{equation}
where $\phi^*=\left(\phi^*_1,\dots,\phi^*_L\right)$ is a vector of linear coefficients to be determined. For brevity, we use the notation $\mathcal{D}(u, \bfx)\phi^*$ for the left-hand side in \eqref{eqn:law2}.
\begin{table*}[]
    \centering
    \begin{tabular}{|c|c|c|}
    \hline
        Name & Equation & Dictionary\\ \hline
       Wave  & $u_{tt} = u_{xx}$ &  $u_{tt}, u_{xx},  u_t, u_x, u,  u^2, uu_x, uu_t$ \\ \hline
       Helmholtz & $(\nabla^2 + k^2)u = 0 $ & $u_{x_0x_0}, u_{x_1x_1} , u_{x_0} , u_{x_1} , u , u^2 , uu_{x_1} , uu_{x_0} $\\ \hline
       Inviscid & $u_t + u_xu = 0 $ & $u_{tt},u_{xx},u_t,u_x,u,u^2,uu_x,u_{xx}^2$ \\ \hline
       KdV & $u_t + u_{xxx} - 6uu_x =0$ & $u_{xxx},u_{tt},u_{xx},u_t,u_x,u,uu_x,u_x^2$ \\ \hline
       Vortex & $u_t + xu_y = yu_x$ & $ u_t, u_x, u_y, x u_x, y u_x, x u_y, y u_y, u$ \\ \hline
       HJB & $u_{x_0x_0} = u_{x_1} + u^2 + u_{x_0}^2$ & $u_{x_1x_1}, u_{x_0x_0},  u_{x_1}, u_{x_0}, u ,u^2, uu_{x_0}, u_{x_0}^2$  \\ \hline
    \end{tabular}
    
    \caption{List of equations and dictionaries used.}
    \label{tab:equations}
\end{table*}
\vspace{-.1cm}

\subsection{Wave Equation Illustration}
As a motivating example, consider the 1D wave equation. 
For brevity, denote by $u_* = \frac{\partial u}{\partial *}$, where $*$ is any variable within the domain of $u$.
Then $u : \Omega \to \bbR$, where $\Omega\subset \bbR^2$, is a solution to the 1D wave equation if
\begin{equation}
u_{tt}-u_{xx} =F(u_{tt},u_{xx}) = 0 \quad \forall x\in \Omega .
\end{equation}
If we then define our dictionary as $\mathcal{D}_1 = u_{tt}$ and $\mathcal{D}_2 = u_{xx}$, then $\mathcal{D}_1(u,x)-\mathcal{D}_2(u,x)=0$ and the PDE is of the form \eqref{eqn:law2} with $\phi^* = (1,-1)$.

\section{Methods}
\label{sec:methods}
Suppose we observe $(\mathbf{x},y)$ defined as in \eqref{eq:noisydatadefinition}, where $u$ is a solution to \eqref{eqn:law2}. Our approach is to approximate $u$ by a neural network function $\hat u :\Omega\to\bbR$, where $\bfx$ is the input/training data. The loss function contains two main components, $\mathcal{L}_u$ and $\mathcal{L}_d$, which we explain in the following subsections.
\vspace{-.1cm}

\subsection{Fitting the Data}
The first component of the loss function is a mean square error (MSE) loss between the observed data points $\{(\bfx_i,y_i)\}_{i=1}^J$ and the value of the neural network at these points. That is, 
\begin{equation}
    \mathcal{L}_u(\bfx;\theta) =\frac{1}{N}\sum_{i=1}^J (y_i-\hat u(\bfx_i,\sigma))^2,
\end{equation}
where $\theta$ are the neural network parameters.
As a consequence of universal approximation theorems \cite{hornik1991approximation}, not only can $u$ be approximated with arbitrary accuracy, by a neural network $\hat u$, but also its derivatives are approximated by $\hat u$ up to arbitrary accuracy.
More formally, if we define the Sobolev norm as 
\begin{equation}
    \|u\|_{m,2} = \sqrt{\sum_{|\alpha|\le m} \|D^{\alpha} u\|^2_2},
\end{equation}
then for any $\epsilon>0$ there exists a neural network $\hat{u}$ such that $|| \hat{u} - u||_{m,2} < \epsilon$. 
Since $u$ is a solution to \eqref{eqn:law2} and $\hat u$ is arbitrarily close to $u$ in Sobolev norm, $\hat u$ is an approximate solution to the same PDE, and the parameters $\{\phi^*_i\}$ can be determined from $\hat u$. 
However, the sampled data contains noise, and minimizing $\mathcal{L}_u$ may lead to $\hat{u}$ overfitting.
We circumvent this by introducing a regularization term. 
\vspace{-.1cm}

\subsection{PDE Estimation}
To estimate the underlying PDE, we first construct a dictionary with $L$ terms, $\mathcal{D} = \{\mathcal{D}_1,\mathcal{D}_2,\ldots,\mathcal{D}_L\}$ and evaluate these functions at $K$ points   $\bfx' = \{\mathbf{x}_i’\}_{i=1}^K$, sampled from $\Omega$, obtaining a $\mathbb{R}^{K \times L}$ matrix with entries 
$$\mathcal{D}(\hat u, \bfx')_{kl} := \mathcal{D}_l(\bfx'_k,\hat{u} (\bfx'_k),D^\alpha\hat{u}(\bfx'_k)).$$
From \eqref{eqn:law2}, $\mathcal{D}(u,\bfx')\phi^* = 0$, and therefore $\phi^*$ lies in the null space of $\mathcal{D}(u,\bfx')$. Equivalently, $\phi^*$ is a singular vector of $\mathcal{D}(u,\bfx')$, with associated singular value $0$. If $|| \hat{u} - u||_{m,2} < \epsilon$, we have, assuming some regularity conditions on $\mathcal D$, that
$$\|\mathcal{D}(u,\bfx')-\mathcal{D}(\hat u,\bfx')\|<C \epsilon,$$
for some constant $C$ that depends on $\mathcal{D}$. Therefore the singular values and associated singular vectors of both matrices are close, and the singular vector of $\mathcal{D}(\hat u,\bfx')$, associated with its smallest singular value, is an approximation of $\phi^*$. In order to calculate the smallest singular vector of $\mathcal{D}(\hat u,\bfx')$, we invoke the min-max theorem for singular values \cite[Theorem~8.6.1]{van1983matrix}, which states the minimum of $||\mathcal{D}(\hat u,\bfx')\phi||$, subject to $||\phi||=1$, is the smallest singular value of $\mathcal{D}(\hat u,\bfx')$.
Therefore, we introduce the loss term
\begin{equation} \label{eq:lddef}
\mathcal{L}_d(\bfx';\theta,\phi)  = \| \mathcal{D}(\hat u(\bfx';\theta),\bfx')\phi \|_2^2,
\end{equation}
and add the constraint $\|\phi\| = 1$. We note that enforcing these constraint is necessary, otherwise the minimizer of \eqref{eq:lddef} would be $\phi = 0$. The contribution of $\mathcal{L}_d$ is twofold: while minimizing \eqref{eq:lddef} over $\phi$ recovers the PDE, minimizing \eqref{eq:lddef} over $\hat u$ forces it to be a solution to the PDE, and prevents it from overfitting to noise.
 
Finally, as we believe that laws of nature are inherently simpler and depend on fewer terms, we impose an additional $\ell^1$ loss term $\phi$ to promote sparsity in the recovered law.
 \begin{equation}
     \mathcal{L}_\text{sparse} = \|\phi\|_1
 \end{equation}

 \subsection{Combining Losses}
 We combine the different loss terms in the following manner to obtain the final loss function we optimize. 
 \begin{equation}
     \mathcal{L}(\bfx,\bfx';\phi,\theta) = \lambda_u\mathcal{L}_u^{1/2}(1+\lambda_d\mathcal{L}_d + \lambda_\text{sparse}\mathcal{L}_\text{sparse}).
 \end{equation}
 The multiplicative scaling of $\mathcal{L}_u$ on the other loss terms acts as an additional regularizer for the function. When $\mathcal{L}_u$ is large, this means the noise variance is larger, and thus the contribution of the other terms increases in order to force $\hat u$ to be smooth and prevent it from overfitting to noise. On the other hand, when $\mathcal{L}_u$ is small, the noise variance is smaller and it is more important for $\hat u$ to interpolate the data.
% To summarize, our final optimization problem is the following:
% \begin{equation}
% \begin{aligned}
%     & \underset{\theta,\phi}{\text{minimize}} & \mathcal{L}(\phi,\theta) \\
%     & \text{subject to} & \|\phi\|_2 = 1
%     \end{aligned}
% \end{equation}
 
 \begin{table*}[]
\newcommand{\timesten}{\text{\footnotesize $\times10$}}
\centering
\begin{tabular}{ | c | c | c | c |}
\hline
Noise Level & 1  & 0.01 & 0 \\ \hline 
Wave & $5.55\timesten^{-2} \pm 9.23\timesten^{-3}$ & $1.48\timesten^{-3} \pm 5.75\timesten^{-4}$ & $6.06\timesten^{-4} \pm 4.51\timesten^{-4}$ \\  \hline
Helmholtz & $3.60\timesten^{-2} \pm 9.58\timesten^{-3}$ & $1.34\timesten^{-2} \pm 2.19\timesten^{-3}$ & $4.75\timesten^{-3} \pm 2.36\timesten^{-3}$ \\ \hline
Vortex & $7.81\timesten^{-2} \pm 4.14\timesten^{-2}$ & $7.98\timesten^{-4} \pm 4.45\timesten^{-4}$ & $9.41\timesten^{-4} \pm 2.05\timesten^{-4}$ \\ \hline
HJB & $1.98\timesten^{-1} \pm 2.07\timesten^{-1}$ & $2.64\timesten^{-3} \pm 1.31\timesten^{-3}$ & $2.42\timesten^{-3} \pm 1.27\timesten^{-3}$ \\ \hline
Inviscid & $5.65\timesten^{-1} \pm 3.77\timesten^{-1}$ & $5.78\timesten^{-4} \pm 0.00$ & $1.85\timesten^{-4} \pm 1.69\timesten^{-4}$ \\ \hline
KdV & $6.84\timesten^{-1} \pm 4.40\timesten^{-1}$ & $6.50\timesten^{-1} \pm 4.59\timesten^{-1}$ & $2.15\timesten^{-2} \pm 8.47\timesten^{-3}$ \\ \hline
\end{tabular}
\caption{Table showing average and standard deviation of error for different noise levels and equations. Here the noise level represents $\sigma$ in \eqref{eq:noisydatadefinition} and the error is defined by \eqref{eq:error}.}
\label{tab:results}
\end{table*}
 
 \section{A Cramer-Rao bound for a simple ODE}
 \label{sec:cramerrao} 
 
 In this section we present a lower bound on estimating a simple ordinary differential equation (ODE), using the Cramér-Rao bound \cite{cramer1946mathematical}, in order to assess the performance of our method. We consider the ODE
 \begin{equation}\label{eq:ODE}
 \frac{du}{dt} = a u.
 \end{equation}
 Our goal is to estimate $a\in \bbR$ by observing noisy data points of a solution to the ODE. All solutions of this ODE are of the form
 \begin{equation}\label{eq:ODEudef}
 u(t)=C e^{a t},
 \end{equation}
 for some latent variable $C\in \bbR$. Consider we observe $J$ data points $ \{(t_1,y_1) \dots  , (t_J,y_J)\}$, with each $t_i$ drawn independently from the uniform distribution in $[0,1]$, $y_i$ defined as in \eqref{eq:noisydatadefinition}, and $u(t_i)$ defined by \eqref{eq:ODEudef}. The probability distribution of $(t_i,y_i)$ is given by:
 \begin{align}
 \nonumber f(y_i,t_i;a,C) &= f(y_i|t_i;a,C) f(t_i),\\
 \label{eq:CRprobdistdef}&= \frac1{\sqrt{2\pi} \sigma} \exp\left(
 \frac{\left(y_i-C \exp(at_i)\right)^2}{2\sigma^2}\right),
 \end{align}
 where we used $f(t_i)=1$, since $t_i$ is drawn from the uniform distribution in $[0,1]$, and $y_i-u(t_i)$ is a centered Gaussian variable with variance $\sigma^2$. If $\widehat a$ and $\widehat C$ are unbiased estimators for $a$ and $C$, respectively, the Cramer-Rao bound \cite{cramer1946mathematical} states that
 \begin{equation}\label{eq:CramerRao}
 \operatorname{Cov}\left(\left[\begin{array}{c}
 \widehat  a \\ \widehat C
 \end{array}\right]\right) \succeq \frac{1}{J} F^{-1} = \frac{1}{J}\left[\begin{array}{cc}
 F_{aa} & F_{aC} \\ F_{aC} & F_{CC}
 \end{array}\right]^{-1},
 \end{equation}
 where $F$ is the Fisher information matrix, $J$ is the number of data points and we write $A\succeq B$ if $A-B$ is a positive semi-definite matrix. The entries of $F$ are defined by
 \begin{equation}\label{eq:Fdef}
 F_{pq} = \int_0^1 \int_{-\infty}^{\infty} \frac{\frac{\partial f}{\partial p} (y,t;a,C) \frac{\partial f}{\partial q} (y,t;a,C)}{f(y,t;a,C)} \,dy\,dt,
 \end{equation}
 where $(p,q)$ is any of $(a,a)$, $(a,C)$ or $(C,C)$, and $f$ is defined in \eqref{eq:CRprobdistdef}. If we define $\text{MSE}:=\bbE[(\widehat{a} - a)^2]$, or equivalently $\text{MSE}=\operatorname{Var}(\widehat{a})$, since $\widehat{a}$ is an unbiased estimator, \eqref{eq:CramerRao} implies
 \begin{align}
 \nonumber \text{MSE}&\ge \frac{1}{J}(F^{-1})_{aa}\\
 \label{eq:CramerRao2} &= \frac{1}{J}\frac{F_{CC}}{F_{aa} F_{CC} - F^2_{aC}}.
 \end{align}
 We now use \eqref{eq:Fdef} to evaluate \eqref{eq:CramerRao2}. For brevity, we omit the evaluation, and just present the final expression:
 \begin{equation}\label{eq:MSElowerbound}
 \text{MSE}\ge \frac{8 \sigma^2 }{C^2 J} \frac{ a^3 e^{-a} \sinh a}{\cosh(2a)-1-2a^2}.
 \end{equation}
 
\begin{figure}
     \centering
     \vspace{-.5cm}
     \includegraphics[width=0.8\linewidth]{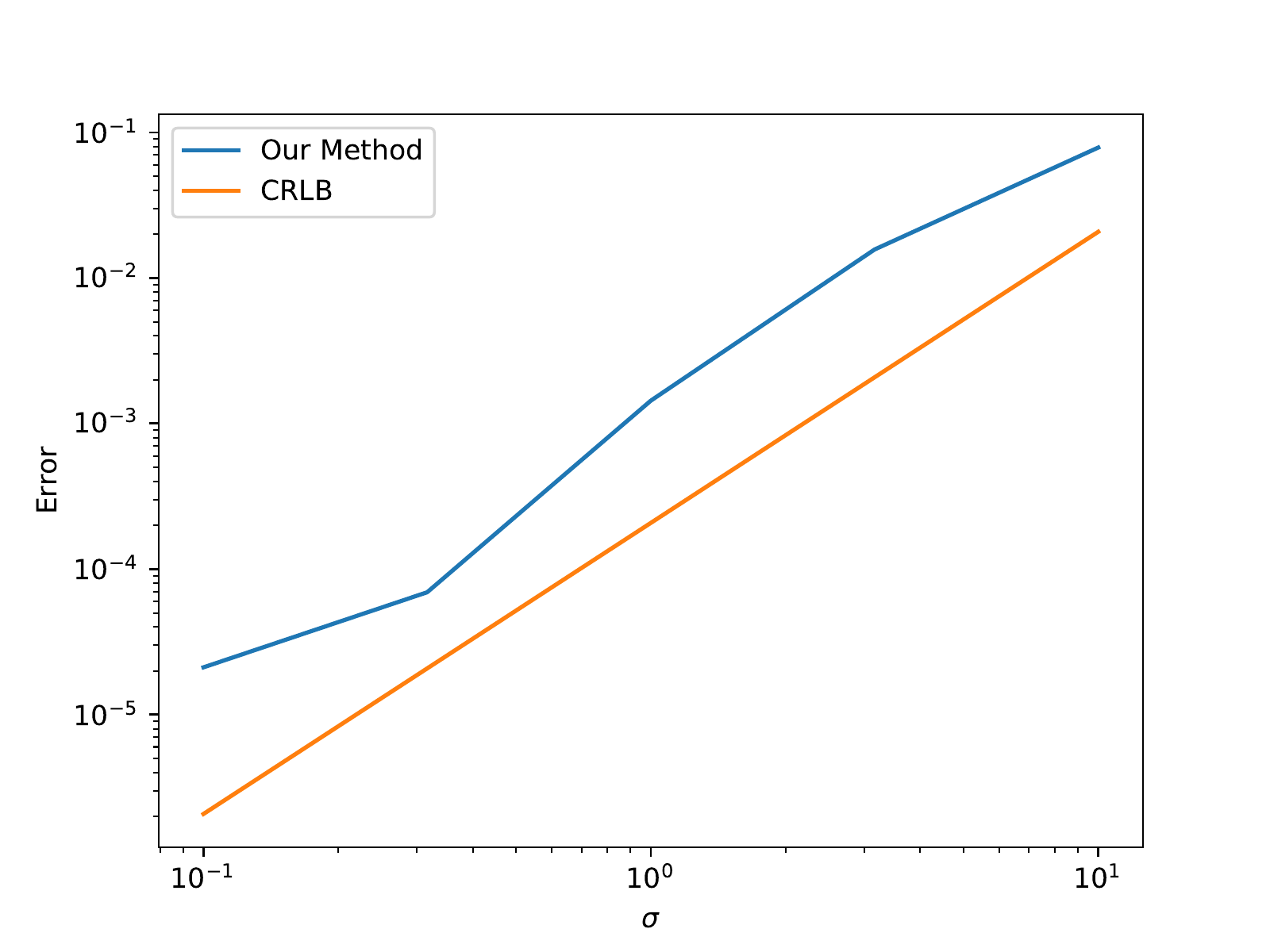}
     \caption{Plot comparing the derived Cramer Rao lower bound (CRLB) to observed empirical results of the described algorithm for 5 noise levels. Algorithms performance is comparable to CRLB shown in orange. }
     \label{fig:crlb}
 \end{figure}

 In order to compare our proposed method to the Cramer-Rao lower bound, we consider two dictionary functions, $\mathcal{D}_1(t) = u(t)$ and $\mathcal{D}_2(t) = u_t(t)$, we generate $1000$ data points as in \eqref{eq:noisydatadefinition}, with $u(t)=e^t$, and check if our algorithm recovers \eqref{eq:ODE} with $a=1$. We run the experiment 3 times and compare the error with \eqref{eq:MSElowerbound}. Finally, we plot the results in Fig.~\ref{fig:crlb}.

\section{Numerical Simulations}
\label{sec:experiments}

As motivating examples, we consider the wave equation, the Helmholtz equation, the Korteweg-de Vries (KdV) equation, a simulated vortex, and Hamilton-Jacobi-Bellman (HJB) equation. A summary of these equations and of the dictionaries used is presented in Table \ref{tab:equations}.
Our neural network is a feed forward fully connected neural network, 4 hidden layers, 50 neurons per hidden layer similar to \cite{berg2019data}, using the softplus \cite{glorot2011deep} activation function as the nonlinearity.
For optimization, we use the Adam optimizer with learning rate of $0.02$ for the parameter $\phi$ and $0.002$ for $\theta$. 
The learning rate for both decays exponentially by $0.9998$ each epoch.
All hyperparameters are maintained constant through all different experiments.
Finally, we sample 10000 points randomly within the domain for the function approximation.

In order to measure the error of our PDE recovery, we use the following formula
\begin{equation}
    \text{err}(\phi,\phi^*) = \sqrt{1 - \Bigg|\frac{\phi^\text{T}\phi^*}{\|\phi\|_2\|\phi^*\|_2}\Bigg|}.
    \label{eq:error}
\end{equation}
We note this quantity is always non-negative, and is $0$ if and only if $\phi$ and $\phi^*$ are colinear. Moreover, this formula also gives an estimate on how many digits of the recovered coefficients are correct: if the error is of the order of $10^{-n}$, then $\phi$ has $n$ correct digits. 

For all the equations in Table \ref{tab:equations} and noise levels $\sigma=\{1,0.01,0\}$, we run our method 3 times and present the results in Table~ \ref{tab:results}. We include the associated code at \url{https://github.com/alluly/pde-estimation}.

\section{Discussion}
\label{sec:majhead}

We present a method for reconstructing the underlying PDE for a set of data while being robust to noise and generalizing to a variety of PDEs.
We consider the method's applicability to multiple equations at various noise levels.
Finally, we show that the method approaches the theoretical bound for parameter estimation for noisy cases for a simple ODE.
In the future, extensions to stochastic differential equations and higher dimensional equations should be considered.
While we present all results for a fixed set of algorithm hyperparameters, it is unclear if these hyperparameters are optimal, and the effects of different hyperparameter regimes should be studied.
Different algorithm hyperparameters may aid in reducing the variability of the recovered solutions.
The results present an opening to apply our method to problems where the estimation of an underlying PDE is necessary.

% Below is an example of how to insert images. Delete the ``\vspace'' line,
% uncomment the preceding line ``\centerline...'' and replace ``imageX.ps''
% with a suitable PostScript file name.
% -------------------------------------------------------------------------

%\vfill
%\pagebreak

\vfill\pagebreak

\bibliographystyle{IEEEbib}
\bibliography{refs}

\end{document}